\pdfoutput=1

\documentclass{article} 
\usepackage{nips12final_e,times}

\usepackage{amsmath,amsthm,amsfonts,accents}
\usepackage{graphicx}
\usepackage{algorithm}
\usepackage{algorithmic}
\usepackage{mymath}
\usepackage{multirow}

\nipsfinalcopy

\title{Matrix Approximation\\under Local Low-Rank Assumption}

\author{
Joonseok Lee$^\textnormal{a}$, Seungyeon Kim$^\textnormal{a}$, Guy Lebanon$^\textnormal{a, b}$, Yoram Singer$^\textnormal{b}$\\
$^\textnormal{a}$ College of Computing, Georgia Institute of Technology, Atlanta, GA 30332\\
$^\textnormal{b}$ Google Research, Mountain View, CA 94043\\
\texttt{\{jlee716@, seungyeon.kim@, lebanon@cc.\}gatech.edu, singer@google.com} \\
}

\newlength{\dhatheight}
\newcommand{\doublehat}[1]{%
    \settoheight{\dhatheight}{\ensuremath{\hat{#1}}}%
    \addtolength{\dhatheight}{-0.25ex}%
    \hat{\vphantom{\rule{1pt}{\dhatheight}}%
    \smash{\hat{#1}}}}

\begin{document}

\maketitle

\begin{abstract} \vspace{-0.25cm}
Matrix approximation is a common tool in machine learning for
building accurate prediction models for recommendation
systems, text mining, and computer vision. A prevalent assumption in
constructing matrix approximations is that the partially observed
matrix is of low-rank. We propose a new matrix approximation model
where we assume instead that the matrix is only \emph{locally} of
low-rank, leading to a representation of the observed matrix as a
weighted sum of low-rank matrices. We analyze the accuracy of the
proposed local low-rank modeling. Our experiments show improvements
of prediction accuracy in recommendation tasks.
\end{abstract}

\section{Introduction}
\label{sec:intro}

Matrix approximation is a common task in machine learning. Given a
few observed matrix entries $\{M_{a_1,b_1},\ldots,M_{a_m, b_m}\}$,
matrix approximation constructs a matrix $\hat M$ that approximates
$M$ at its unobserved entries. In general, the problem of completing
a matrix $M$ based on a few observed entries is ill-posed, as there
are an infinite number of matrices that perfectly agree with the
observed entries of $M$. Thus, we need additional assumptions such
that $M$ is a low-rank matrix. More formally, we approximate a
matrix $M\in\R^{n_1\times n_2}$ by a rank-$r$ matrix $\hat M =
UV^T$, where $U\in\R^{n_1\times r}$, $V\in\R^{n_2\times r}$, and $r
\ll \min(n_1, n_2)$.
In this note, we assume that $M$ behaves as a low-rank matrix in
the vicinity of certain row-column combinations, instead of assuming
that the entire $M$ is low-rank. We therefore construct several
low-rank approximations of $M$, each being accurate in a particular
region of the matrix. Smoothing the local low-rank approximations,
we express $\hat M$ as a linear combination of low-rank matrices
that approximate the unobserved matrix $M$. This mirrors the theory
of non-parametric kernel smoothing, which is primarily developed for
continuous spaces, and generalizes well-known compressed sensing
results to our setting.

\section{Global and Local Low-Rank Matrix Approximation}
\label{sec:lrma}
We describe in this section two standard approaches for low-rank
matrix approximation (LRMA). The original (partially observed)
matrix is denoted by $M\in\R^{n_1\times n_2}$, and its low-rank
approximation by $\hat{M} = UV^T$, where $U\in\R^{n_1\times r}$,
$V\in\R^{n_2\times r}$, $r \ll \min(n_1, n_2)$.

\vspace{-0.25cm}
\paragraph{Global LRMA} %
Incomplete SVD is a popular approach for constructing a
low-rank approximation $\hat{M}$ by minimizing the Frobenius norm
over the set $\mathtt{A}$ of observed entries of $M$:
\begin{align}\label{eq:regSVD1}
(U,V) = \argmin_{U,V} \sum_{(a,b) \in \mathtt{A}} ([UV^T]_{a,b} -
M_{a,b})^2.
\end{align}
Another popular approach is minimizing the nuclear norm of a matrix
(defined as the sum of singular values of the matrix) satisfying
constraints constructed from the training set:
\begin{align} \label{eq:normRecovery2}
\hat M = \argmin_{X} \|X\|_*, \  \ \text{s.t.}\quad
\|\Pi_{\mathtt{A}}(X - M)\|_F < \alpha ~
\end{align}
where $\Pi_\mathtt{A}: \R^{n_1\times n_2} \to \R^{n_1\times n_2}$ is
the projection defined by $[\Pi_\mathtt{A}(M)]_{a, b} = M_{a, b}$ if $(a, b) \in \mathtt{A}$ and 0 otherwise, and $\|\cdot\|_F$ is the Frobenius norm.

Minimizing the nuclear norm $\|X\|_*$ is an effective surrogate for
minimizing the rank of $X$. One advantage of
\eqref{eq:normRecovery2} over \eqref{eq:regSVD1} is that we do not
need to constrain the rank of $\hat{M}$ in advance. However, problem
\eqref{eq:regSVD1} is substantially easier to solve than problem
\eqref{eq:normRecovery2}.

\begin{figure}[t]
\centering
\includegraphics[scale=0.385]{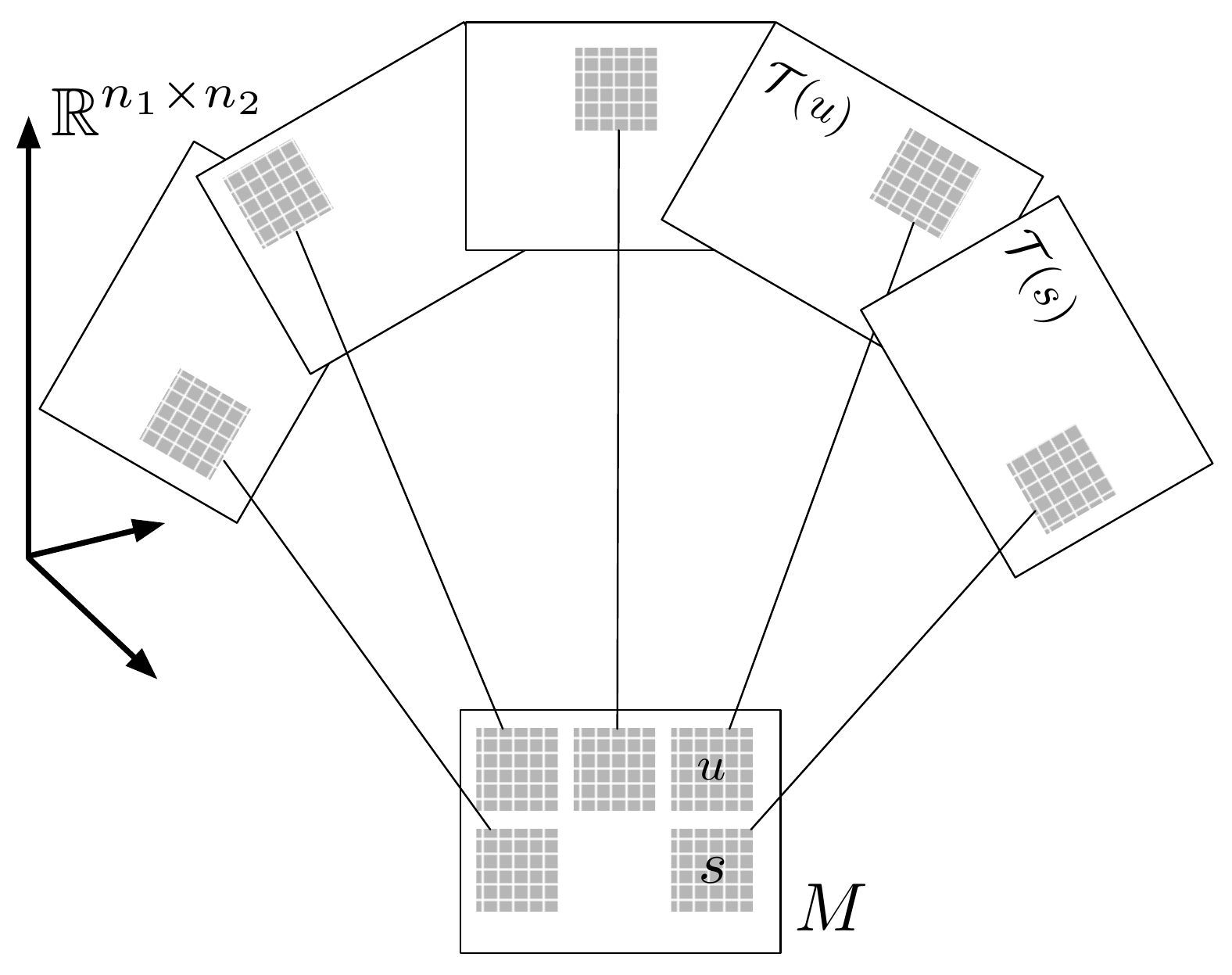}
\includegraphics[scale=0.957]{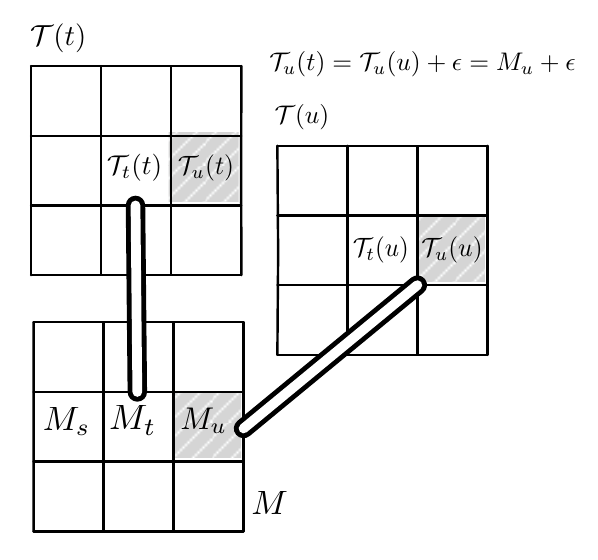}
\caption{For illustrative purposes, we assume a distance function
$d$ whose neighborhood structure coincides with the natural order on
indices. That is, $s=(a,b)$ is similar to $u=(a',b')$ if $|a-a'|$
and $|b-b'|$ are small. \textit{(Left)} For all
$s\in[n_1]\times[n_2]$, the neighborhood $\{s':d(s,s')<h\}$ in the
original matrix $M$ is approximately described by the corresponding
entries of the low-rank matrix $\mathcal{T}(s)$ (shaded regions of
$M$ are matched by lines to the corresponding regions in
$\mathcal{T}(s)$ that approximate them). If $d(s,u)$ is small,
$\mathcal{T}(s)$ is similar to $\mathcal{T}(u)$, as shown by their
spatial closeness in the embedding space $\mathbb{R}^{n_1\times
n_2}$. \textit{(Right)} The original matrix $M$ (bottom) is
described locally by the low-rank matrices $\mathcal{T}(t)$ (near
$t$) and $\mathcal{T}(u)$ (near $u$). The lines connecting the three
matrices identify identical entries: $M_t=\mathcal{T}_t(t)$ and
$M_u=\mathcal{T}_u(u)$. The equation at the top right shows a
relation tying the three patterned entries. Assuming the distance
$d(t,u)$ is small,
$\epsilon=\mathcal{T}_u(t)-\mathcal{T}_u(u)=\mathcal{T}_u(t)-M_u(u)$
is small as well.} \label{fig:model}
\end{figure}

\vspace{-0.25cm}
\paragraph{Local LRMA}
In order to facilitate a local low-rank matrix approximation, we
need to pose an assumption that there exists a metric structure over
$[n_1]\times[n_2]$, where $[n]$ denotes the set of integers $\{ 1,
\dots, n\}$. Formally, $d((a,b),(a',b'))$ reflects the similarity
between the rows $a$ and $a'$ and columns $b$ and $b'$.
In the global matrix factorization setting above, we assume that the
matrix $M\in\mathbb{R}^{n_1\times n_2}$ has a low-rank structure. In
the local setting, however, we assume that the model is
characterized by multiple low-rank $n_1\times n_2$ matrices.
Specifically, we assume a mapping
$\mathcal{T}:[n_1]\times[n_2]\to\mathbb{R}^{n_1\times n_2}$ that
associates with each row-column combination $[n_1]\times[n_2]$ a low
rank matrix that describes the entries of $M$ in its neighborhood
(in particular this applies to the observed entries $\mathtt{A}$): $
\mathcal{T}:[n_1]\times[n_2]\to\mathbb{R}^{n_1\times n_2}$ where
$\mathcal{T}_{a,b}(a,b) = M_{a,b}$. Note that in contrast to the
global estimate in Global LRMA, our model now consists of multiple
low-rank matrices, each describing the original matrix $M$ in a
particular neighborhood. Figure~\ref{fig:model} illustrates this
model.

Without additional assumptions, it is impossible to estimate the
mapping $\mathcal{T}$ from a set of $m< n_1n_2$ observations. Our
additional assumption is that the mapping $\mathcal{T}$ is slowly
varying. Since the domain of $\mathcal{T}$ is discrete, we assume
that $\mathcal{T}$ is H\"older continuous. Following common
approaches in non-parametric statistics, we define a smoothing
kernel $K_h(s_1,s_2)$, where $s_1,s_2\in [n_1]\times[n_2]$, as a
non-negative symmetric unimodal function that is parameterized by a
bandwidth parameter $h>0$. A large value of $h$ implies that
$K_h(s,\cdot)$ has a wide spread, while a small $h$ corresponds to
narrow spread of $K_h(s,\cdot)$. We use, for example, the
Epanechnikov kernel, defined as $K_h(s_1,s_2) = \frac{3}{4} (1 -
d(s_1, s_2)^2) \textbf{1}_{\{d(s_1,s_2)< h\}}$. We denote by
$K^{(a,b)}_h$ the matrix whose $(i,j)$-entry is $K_h((a,b),(i,j))$.

Incomplete SVD \eqref{eq:regSVD1} and compressed sensing
\eqref{eq:normRecovery2} can be extended to local version
as follows
\begin{align}
\text{Incomplete SVD:} & \ \ \hat{\mathcal{T}}(a,b) = \argmin_{X} \|
K^{(a,b)}_h\odot{\Pi}_{\mathtt{A}} (X - M)\|_F \ \
 \; \text{s.t.}\quad  \text{rank}(X)=r\label{eq:regSVDLocal2}\\
\text{Compressed Sensing:} & \ \ \hat{\mathcal{T}}(a,b) =
\argmin_{X} \|X\|_* \ \ \; \text{s.t.}\quad
\|K^{(a,b)}_h\odot{\Pi}_{\mathtt{A}}(X - M)\|_F <
\alpha, \label{eq:normRecoveryLocal2}
\end{align}
where $\odot$ denotes a component-wise product of two matrices,
$[A\odot B]_{i,j}=A_{i,j}B_{i,j}$.

The two optimization problems above describe how to estimate
$\hat{\mathcal{T}}(a,b)$ for a particular choice of $(a,b)\in
[n_1]\times[n_2]$. Conceptually, this technique can be applied for
each test entry $(a,b)$, resulting in the matrix approximation
$\hat{M}_{a,b} = \hat{\mathcal{T}}_{a,b}(a,b)$, where
$(a,b)\in[n_1]\times[n_2]$. However, this requires solving a
non-linear optimization problem for each test index $(a,b)$ and is
thus computationally prohibitive. Instead, we use Nadaraya-Watson
local regression with a set of $q$ local estimates
$\hat{\mathcal{T}}(s_1), \ldots, \hat{\mathcal{T}}(s_q)$, in order
to obtain a computationally efficient estimate
$\doublehat{\mathcal{T}}(s)$ for all $s\in [n_1]\times[n_2]$:
\begin{align} \label{eq:NWRegression}
\doublehat{\mathcal{T}}(s) = \sum_{i=1}^q \frac{ K_h(s_i,s)}{\sum_{j=1}^q
K_h(s_{j},s)} \,\hat{\mathcal{T}}(s_i) ~ .
\end{align}
Equation \eqref{eq:NWRegression} is simply a weighted average of
$\hat{\mathcal{T}}(s_1), \ldots, \hat{\mathcal{T}}(s_q)$, where the
weights ensure that values of $\hat{\mathcal{T}}$ at indices close
to $s$ contribute more than indices further away from $s$.

Note that the local version can be faster than global SVD since (a)
each low-rank approximation is independent of each other, so can be
computed in parallel, and (b) the rank used in the local SVD model
can be significantly lower than the rank used in a global one. If
the kernel $K_h$ has limited support ($K_h(s,s')$ is often zero),
the regularized SVD problems would be sparser than the global SVD
problem, resulting in additional speedup.





\section{Experiments}
\label{sec:exp}

\begin{figure*}[t]
\centering
\includegraphics[scale=0.45]{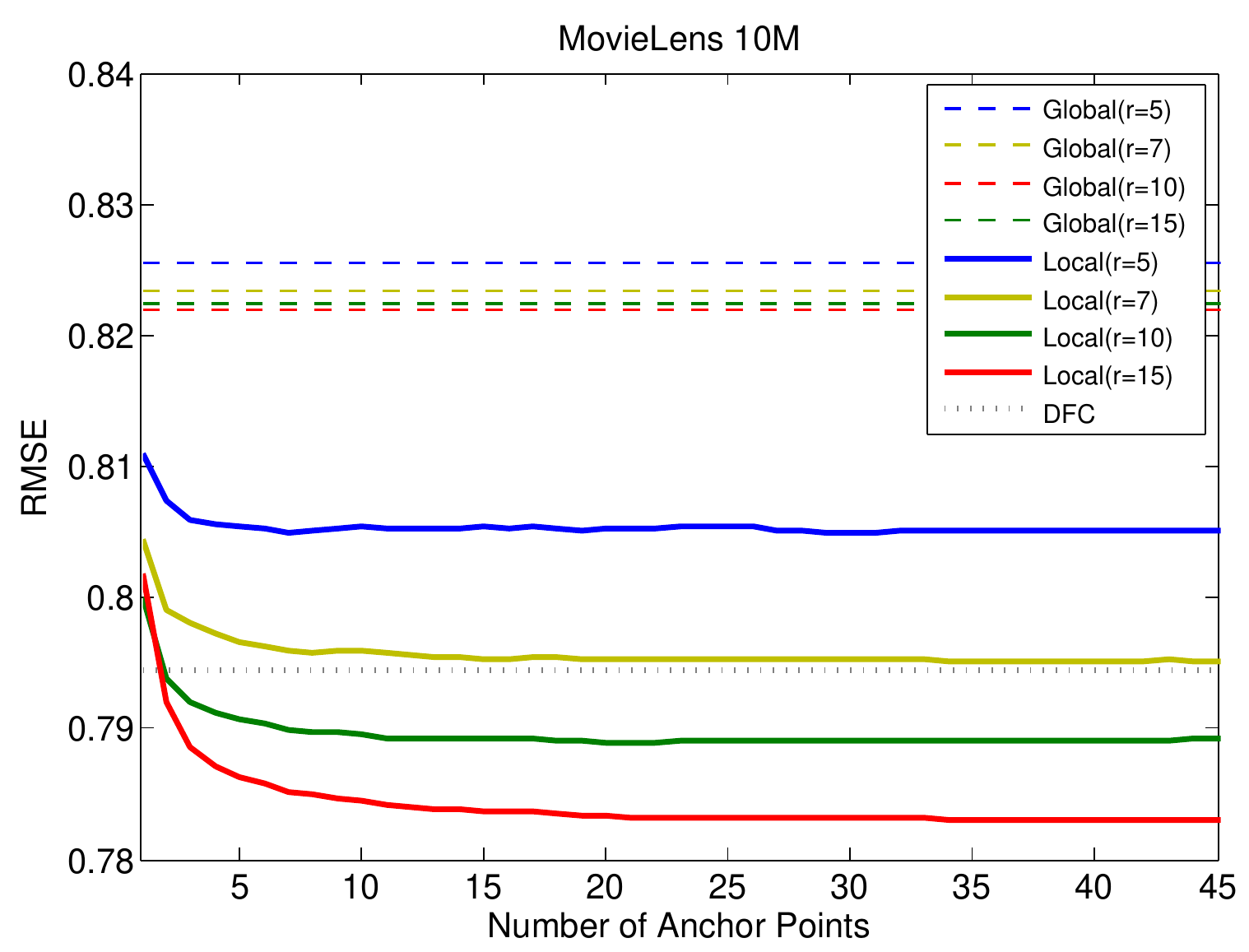}
\includegraphics[scale=0.45]{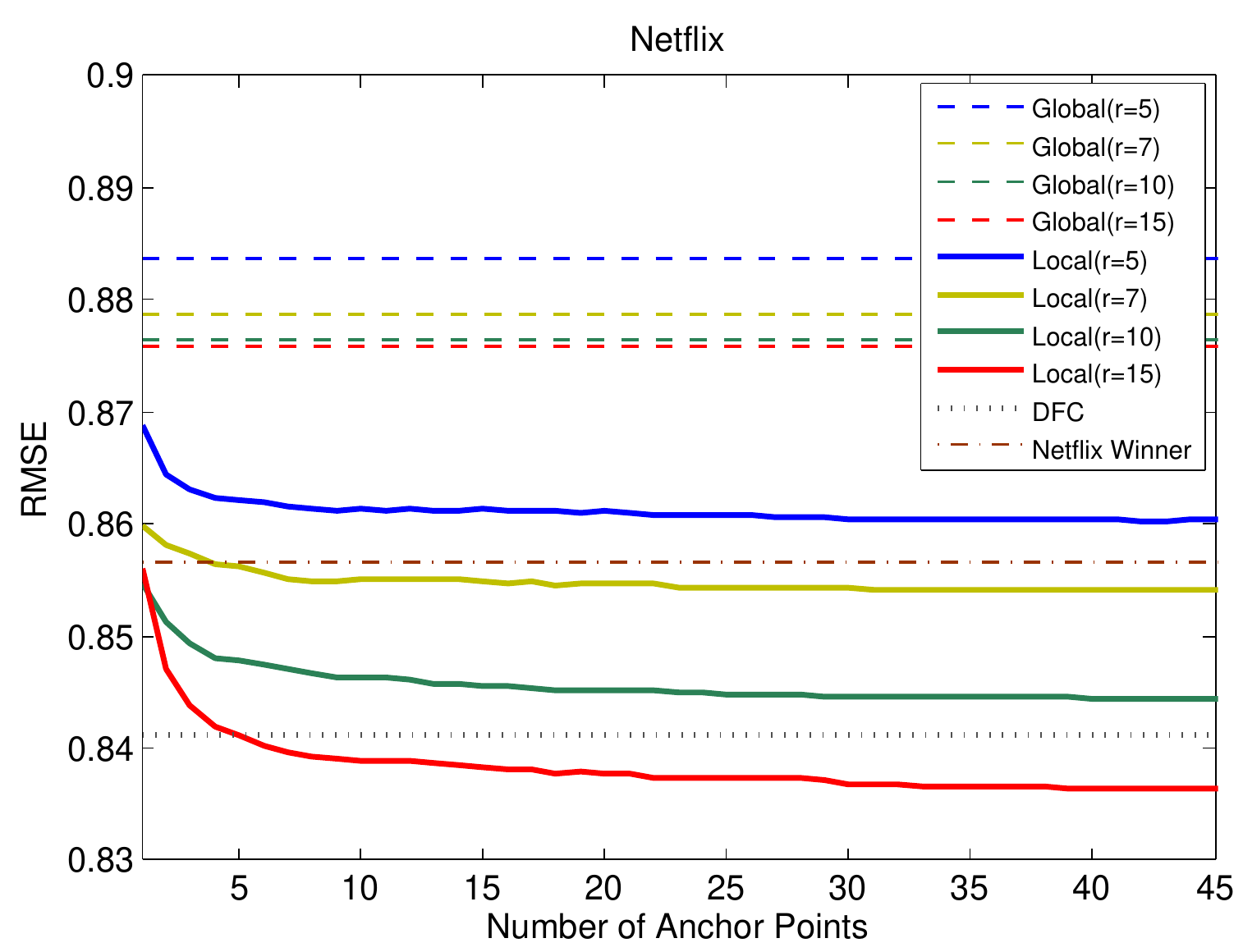}
\caption{RMSE of global-LRMA, local-LRMA, and other baselines on
MovieLens 10M \textit{(Left)} and Netflix \textit{(Right)} dataset.
Local-LRMA models are indicated by thick solid lines, while
global-LRMA models are indicated by dotted lines. Models with same
rank are colored identically.} \label{fig:trend}
\end{figure*}
We compare local-LRMA to global-LRMA and other state-of-the-art
techniques on popular recommendation systems datasets: MovieLens 10M
and Netflix. We split the data into 9:1 ratio of train and test set.
A default prediction value of 3.0 was used whenever we encounter a
test user or item without training observations. We use the
Epanechnikov kernel with $h_1=h_2=0.8$, assuming a product form
$K_h((a, b), (c, d)) = K_{h_1}'(a, c) K_{h_2}''(b, d)$. For distance
function $d$, we use arccos distance, defined as $d(x, y) = \arccos
\left( \langle x, y\rangle / \|x\| \|y\| \right)$. Anchor points
were chosen randomly among observed training entries. $L_2$
regularization is used for local low-rank approximation.

Figure~\ref{fig:trend} graphs the RMSE of Local-LRMA and global-LRMA
as well as the recently proposed method called DFC
(Divide-and-Conquer Matrix Factorization) as a function of the
number of anchor points. Both local-LRMA and global-LRMA improve as
$r$ increases, but local-LRMA with rank $r\geq 5$ outperforms
global-LRMA with any rank. Moreover, local-LRMA outperforms
global-LRMA in average with even a few anchor points (though the
performance of local-LRMA improves further as the number of anchor
points $q$ increases).

\end{document}